# A Noncontact Technique for Wave Measurement Based on Thermal Stereography and Deep Learning

Deyu Li, Longfei Xiao, Handi Wei, *Member, IEEE*, Yan Li, and Binghua Zhang

*Abstract*—The accurate measurement of the wave field and its spatiotemporal evolution is essential in many hydrodynamic experiments and engineering applications. The binocular stereo imaging technique has been widely used to measure waves. However, the optical properties of indoor water surfaces, including transparency, specular reflection, and texture absence, pose challenges for image processing and stereo reconstruction. This study proposed a novel technique that combined thermal stereography and deep learning to achieve fully noncontact wave measurements. The optical imaging properties of water in the long-wave infrared spectrum were found to be suitable for stereo matching, effectively avoiding the issues in the visible-light spectrum. After capturing wave images using thermal stereo cameras, a reconstruction strategy involving deep learning techniques was proposed to improve stereo matching performance. A generative approach was employed to synthesize a dataset with ground-truth disparity from unannotated infrared images. This dataset was then fed to a pretrained stereo neural network for fine-tuning to achieve domain adaptation. Wave flume experiments were conducted to validate the feasibility and accuracy of the proposed technique. The final reconstruction results indicated great agreement and high accuracy with a mean bias of less than 2.1% compared with the measurements obtained using wave probes, suggesting that the novel technique effectively measures the spatiotemporal distribution of wave surface in hydrodynamic experiments.

## I. Introduction

WAVE flume experiment is a crucial foundation for researching wave hydrodynamics, providing validation for theory research and numerical computations. Accurately measuring spatiotemporal wave fields has been a challenging task in the laboratory, especially for the near-field wave of the structure in extreme wave-structure interactions. The invasive wave probe [1] has been the standard instrument in the laboratory due to its precision and low cost. However, the wave probe can only perform single-point measurements and inevitably disturbs the flow field. Especially in limited areas, such as measuring wave runup near walls, deploying multiple invasive wave gauges can have a significant impact on the dynamics of transient wave transformation, leading to considerable errors. Therefore, a new measurement technique is necessary to meet the demands for three-dimensional noncontact wave measurements.

As an advanced technique well suited for surface measurement and reconstruction, stereo photogrammetry has been successfully employed in the observation of ocean waves [2], [3], [4]. The prerequisite for binocular stereo reconstruction is that the measured surface behaves Lambertian. Under the influence of winds, waves, and currents at sea, the water surface exhibits rich textural features, and the wave surface can be effectively considered a Lambertian surface when observed from a significant distance [5]. However, the application of this technology to wave field measurements in indoor water tanks or basins presents numerous challenges. Even in cases where the cameras were positioned sufficiently far away from the water surface and there was wind assistance, the direct use of binocular stereo vision method yielded unsatisfactory results [6], [7].

In close-range observations, the optical properties of the water surface, including transparency, specular reflection, and texture loss, pose challenges for image processing and stereo reconstruction. However, the error in binocular stereo measurement is directly proportional to the square of the measured distance [8], and the precision decreases as the distance to the object increases. Thus, to ensure adequate spatial resolution, achieving close-range stereo measurement of the indoor water surface is particularly meaningful for fine experimental studies in fluid mechanics, which can become an effective tool for further studying spatiotemporal wave phenomena.

In some cases, these challenges can be addressed by introducing artificial features to mark the water surface. The methods can be categorized into two main types: techniques based on floating markers and techniques based on the projection of patterns. For the floating markers, available options included wooden particles [9], polypropylene beads [10], foam particles [11], and fluorescent flakes [12]. Although these floating markers provided Lambertian features for the water surface, they were susceptible to aggregation under wave transport, resulting in the loss of image texture and suboptimal reconstruction outcomes. In an attempt to mitigate this issue, the

This work is supported by National Natural Science Foundation of China under Grants 52031006 and 42206192, Natural Science Foundation of Hainan Province under Grant 521QN275, a Research Project of Shanghai Jiao Tong University under Grant SSZX22002. *(Corresponding author: Handi Wei.)*

Deyu Li, Longfei Xiao, Handi Wei, and Yan Li are with the State Key Laboratory of Ocean Engineering, Shanghai Jiao Tong University, Shanghai 200240, China, and also with the SJTU Yazhou Bay Institute of Deepsea Sci-Tech, Sanya 572000, China (e-mail: lideyu@sjtu.edu.cn; xiaolf@sjtu.edu.cn; weihandi@sjtu.edu.cn; liyan97@sjtu.edu.cn).

Binghua Zhang is with Mairun (Shanghai) Intelligent Technology Company Ltd., Shanghai 201201, China (ben.zhang@marautec.com).

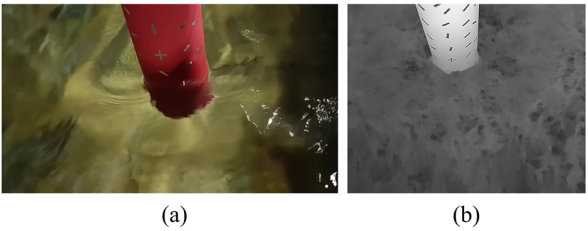

(a)          (b)

Fig. 1. Images of water waves at the visible–light spectrum (a) and infrared spectrum (b).

floating markers were interconnected into a marker-net [13], [14]. However, this approach inevitably affected the surface hydrodynamic characteristics and proved challenging for measuring water surfaces with large deformations, such as violent wave fields around structures. Due to the low reflectance of the water surface, the techniques based on the projection of diffusive patterns often required specific dyes [15]. The selected dye should offer high reflectivity without affecting the hydrodynamic properties of the waves [16], such as titanium dioxide [17]. In addition, methods based on refraction [18] and specular reflection [19] were also used to deduce the shape of the water surface, but imposed relatively stringent requirements on the measurement environment. However, these methods are impotent when the experimental facility is large or when measuring near-field waves in structures.

These methods attempt to address the issues in wave field measurement in the visible-light spectrum, in which the water surface is transparent and specular, presenting challenges for optical image processing. However, in the infrared spectrum, particularly the long-wave infrared spectrum (8–14 μm), which is the operating range for common thermal imagers, the optical properties of the water surface are significantly different, as presented in Fig. 1. (Unless otherwise specified in the following descriptions, infrared imaging in this paper refers to long-wave infrared imaging and thermal imaging.) In infrared imaging, the water surface becomes opaque, and its infrared reflectance is extremely low, akin to a Lambertian surface. This aligns with the prerequisites of most image processing techniques. In fact, numerous studies have employed infrared imaging to investigate various water wave phenomena, such as the dynamic effects of surfactants on a wind-driven wave surface [20] and the velocity measurement of microscale breaking waves [21].

In a quiescent state, there is heat exchange between the water surface and the air, resulting in temperature variations in the water. Consequently, a thin, cold/warm layer exists on the water surface. When subjected to wave disturbances, mixing of colder and warmer fluids will occur, and subtle temperature differences can be captured using highly sensitive infrared cameras to constitute rich texture features in thermal imaging. This is well-suited for performing stereo matching. The use of infrared thermal imaging cameras to capture wave images eliminates the need for cumbersome operations such as adding markers to the water surface. Furthermore, this approach can fully leverage the natural texture features resulting from thermal exchange, making it highly practical and applicable.

While infrared imaging addresses the problem of wave image acquisition, infrared images, distinct from visible–light images, have some inherent limitations. These images typically exhibit a monochromatic color space, tend to contain more noise, and may have blurred edges. To obtain better reconstruction results, further improvements are required in the stereo matching algorithms. A superior algorithm can compensate for the quality deficiencies inherent in the images. At present, the algorithms applied to stereo matching in wave images mainly include semi-global stereo matching [22] and digital image correlation [12]. These algorithms are based on traditional computer vision principles, relying on the definition of a mathematical optimization function to accomplish stereo matching. However, they are often susceptible to the influence of image noise and quality.

In recent years, stereo matching algorithms based on deep learning techniques, such as PSMNet [23], CREStereo [24], and RAFT-Stereo [25], have achieved superior performances. These algorithms are usually supervised learning, and there are mainly two types of supervised stereo networks [26]: one based on 2D or 3D convolutional neural networks and the other using recurrent neural networks for multiple iterations of disparity optimization. Furthermore, deep-learning-based stereo methods have been employed to address practical challenges in encountered wave field measurement for navigating surface vehicles [27].

However, one significant challenge in applying supervised learning to real-world scenarios is obtaining ground-truth disparity. To address this challenge, two main solutions have emerged over the years. The first was the advent of self-supervised methods [28], [29], which required a large amount of data. However, at present, their accuracy was not distinctly evident. They have proven effective only in specializing or adapting to single domains and often lack generalization [30]. The second solution involved the use of virtual rendering synthetic data for training [31], [32], but a domain gap always exists between the synthetic data and the real world [33].

A highly practical and cost-effective approach involves pretraining a model on a synthetic dataset and then fine-tuning it with a small amount of real-world data to achieve domain adaptation. There are several ways to conduct fine-tuning. For example, one designed a self-supervised loss function [34], but the training often faced the risk of divergence. Another approach employed generative adversarial networks to transfer the style of real-world images into the synthetic images with ground truth [35], [36], followed by training. This was an indirect method, and its efficacy in domain adaptation was limited. In some studies [37], [38], large models, such as monocular depth estimation models or neural radiance fields, have been used to synthesize stereo pairs and disparity from single or a few real-world images. The synthesized data was then fed into the selected stereo neural network for training, achieving direct adaptation to real-world datasets with favorable outcomes. Furthermore, with the continuous optimization of large models and the development of stereo matching algorithms, this training strategy was expected to improve "for free."

In this study, a novel technique was developed based on thermal stereography and a fine-tuned stereo matching neural network to measure the indoor wave surface. A binocular system comprising two infrared thermal imaging cameras with identical parameters was used to acquire textural infrared images of the water surface, solving the problem of specular reflection, transparency, and texture loss in the visible spectrum. Subsequently, a stereo reconstruction strategy was employed based on the deep learning techniques. Leveraging an off-the-shelf monocular depth estimation model, the stereo training data was synthesized from single left images. The synthesized dataset was then fed into a pretrained stereo neural network for fine-tuning, achieving domain adaptation to infrared images. The adapted stereo model had enhanced stereo matching performance on the target dataset, yielding high-quality disparity maps. Finally, wave flume experiments of regular waves and wave–cylinder interactions were performed to validate the feasibility and accuracy of the proposed technique. In particular, the ability to measure near-field waves, which has been one of the measurement difficulties in hydrodynamic experiments, was examined in the wave-cylinder interaction experiment. The acquired infrared image sequences were reconstructed frame-by-frame. The wave time histories were extracted and compared with the measurements by wave probes to evaluate the performance.

The main contributions of this study are summarized as follows:
1) Introduce a novel experimental method using thermal stereography to acquire stereo images with sufficient textures for specular water surfaces in the laboratory.
2) Provide a feasible domain adaptation strategy for the application of a stereo matching network on unannotated infrared images, and obtain improved stereo matching results.
3) The technique achieves three-dimensional reconstruction based on infrared wave images, enhancing the noncontact measurement of wave fields and providing an effective tool for wave hydrodynamics study. The reconstruction results agree well with wave probe measurements.

## II. Thermal Stereography Technique

Acquisition of high-quality images is a crucial prerequisite for binocular stereovision. The images used for stereo reconstruction should exhibit Lambertian characteristics and possess rich textures.

### A. Binocular Stereography

The essence of binocular stereovision lies in triangulation. Typically, a stereo system consists of two horizontally aligned cameras. Fig. 2 presents a schematic diagram of binocular imaging based on the pinhole camera model [39]. The projection points of point $P$ on the left and right camera planes are denoted as $p_1$ and $p_2$, respectively. Together with the two camera centers, $O_L$ and $O_R$, they form a triangle. The coordinate of the object point $P$ in the camera coordinate system is $(x, y, z)$. Based on the principles of similar triangles, the distance of the object point $P$ from the baseline $B$ is commonly referred to as "*depth*" and denoted as $z$:

$$z = \frac{B \cdot f}{d} \quad (1)$$

where $d$ denotes the difference between $p_1$ and $p_2$ along the $x'$-axis, called disparity.

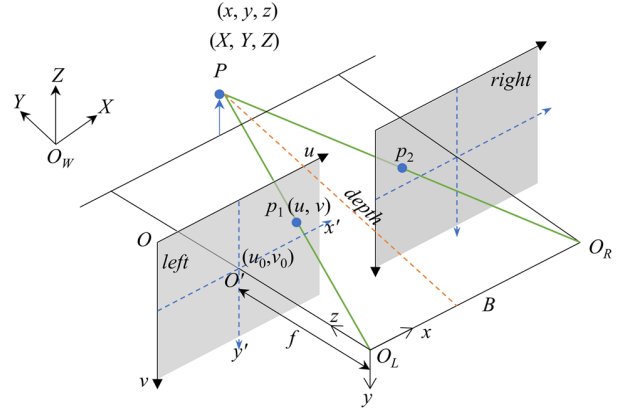

Fig. 2. Schematic diagram of binocular stereo imaging. The measured point $P$ and two optical centers $O_L$ and $O_R$ form a triangle; $p_1$ and $p_2$ denote the projected points on the left and right image plane, respectively; $B$ denotes the distance between two optical centers and is called the baseline; $f$ denotes the focal length; and $xyz$-$O_L$, $x'y'$-$O'$, $uv$-$O$, and $XYZ$-$O_W$ denote the camera, image, pixel, and world coordinate systems, respectively.

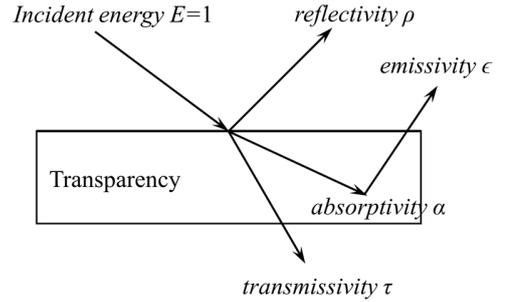

Fig. 3. Different destinations of incident energy for the transparency.

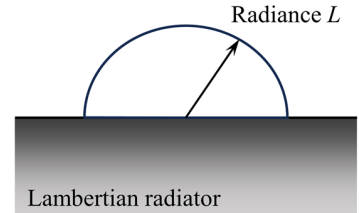

Fig. 4. The Lambertian radiator has constant radiance $L$. The Lambertian surface has the same brightness irrespective of the direction of the observation.

Typically, a stereo matching algorithm is used to obtain a disparity map, from which a point cloud can be generated. The formula for calculating depth $z$ is given, whereas the remaining two coordinates can be determined using (2):

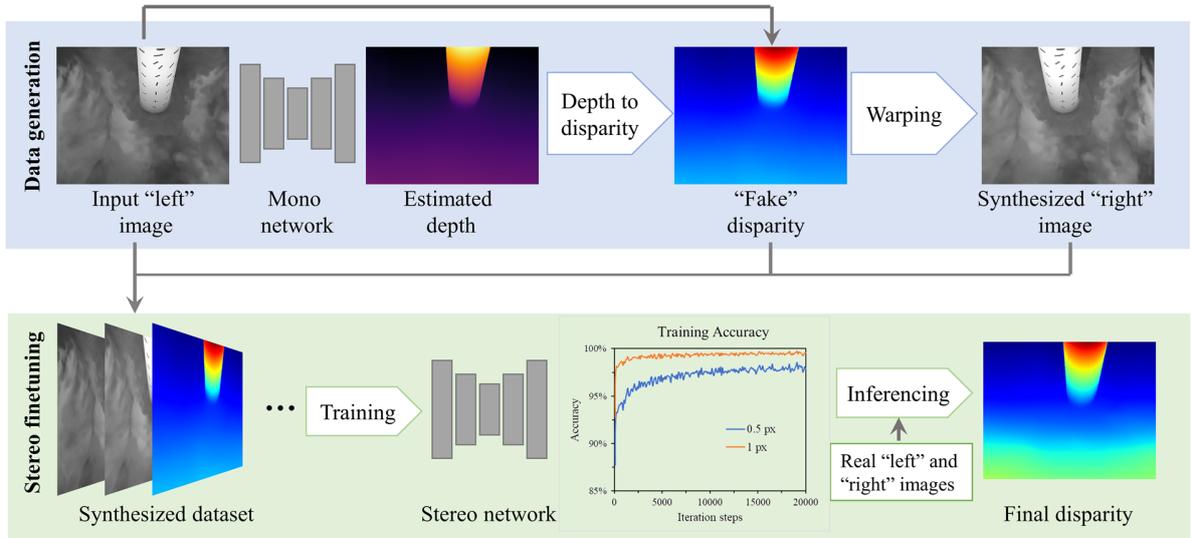

Fig. 5. Overview of the domain adaption approach. Single left image is used to synthesize training data through an off-the-shelf monocular depth estimation network. The synthesized dataset is then fed to a pretrained stereo neural network to reach domain adaptation and obtain fine result in inference. Curves of 0.5 and 1 px represent the training accuracies that disparity errors less than 0.5 and 1 px, respectively.

$$x = \frac{z \cdot (u - u_0)}{f}, y = \frac{z \cdot (v - v_0)}{f} \quad (2)$$

where ($u_0$, $v_0$) denotes the principal point of the image.

*B. Infrared Imaging of Water Surface*

Objects with temperatures above absolute zero emit infrared radiation, and the radiometric effects of the surface properties are described by four parameters: *emissivity ϵ*, *absorptivity α*, *reflectivity ρ*, and *transmissivity τ*. According to Kirchhoff's law [40], when the thermodynamic equilibrium is reached, *absorptivity α* equals *emissivity ϵ*. By conservation of energy, the relationship between these parameters is described in (3) and Fig. 3.

$$\alpha + \rho + \tau = 1 \quad (3)$$

The emitted radiation highly depends on the object's temperature and surface emissivity. In the wavelength range of 8–14 μm, water has a surface emissivity of approximately 95%, implying that the reflectance and transmittance of water in this wavelength range are extremely low. From the perspective of the infrared camera, water is opaque, and specular reflections can be effectively avoided, as presented in Fig. 1. In fact, it is very close to an ideal Lambertian surface with constant radiance $L$, as presented in Fig. 4. The infrared optical properties of the wave surface satisfy the basic premise of binocular stereo vision.

III. STEREO MATCHING BASED ON DEEP LEARNING

The significant differences between infrared images and visible-light images are as follows:
1) Color domain difference: Unlike the CCD/CMOS sensor of visible-light cameras, the infrared detector perceives only the intensity of radiation in the infrared spectrum to generate an image. The color domain of infrared images is singular, providing less information compared with visible images.
2) Noise and blur: Constrained by detector technology, infrared images typically contain more noise. Furthermore, imaging usually depends on the object's temperature and emissivity, resulting in somewhat blurred object boundaries and a lack of distinct variations in brightness.

As a result, traditional stereo matching algorithms struggle when dealing with such infrared images. To more effectively accomplish the reconstruction task of infrared wave images, the deep learning technique was introduced owing to its significant advantages in handling noises and estimating disparities in weak texture and occluded regions.

A deep learning-based stereo matching algorithm named RAFT-Stereo [25] was selected as the baseline method. It was based on an iterative optimization network structure, incorporating lightweight recurrent modules for multiple iterations, showcasing strong cross-domain generalization capabilities and robustness, and demonstrating leading results on some stereo matching evaluation datasets [41], [42]. Although RAFT-Stereo provided pretrained weights on a synthetic dataset, the pretrained model cannot be directly applied due to significant differences between computer-rendered synthetic and real-world images. To bridge this gap, domain adaptation (transfer learning or fine-tuning) was necessary. This involves initializing the network with pretrained weights, conducting a small amount of training on the target dataset, and thus achieving desirable results at a relatively low cost.

*A. Domain Adaptation*

To maintain the stability and efficacy of the stereo model,

| **Algorithm 1** Domain adaptation—Synthesize training data. |
|---|
| **Input:** Target-domain dataset $D_t(I_l, I_r)$ |
| **for** $i \in [0, len(D_t))$ **do** |
|     $L \leftarrow m_d(I_l)$ |
|     Given $[d_{\min}, d_{\max}]$, $\tilde{d}_l \leftarrow L_{Norm} \cdot (d_{\max} - d_{\min}) + d_{\min}$, and calculate occlusion $mask$ |
|     $\tilde{I}_r \leftarrow warp(I_l, \tilde{d}_l, mask)$ |
| **end** |
| **Output:** Synthetic dataset $S_t(I_l, \tilde{I}_r, \tilde{d}_l)$ |

| **Algorithm 2** Calculate occlusion mask. |
|---|
| **Input:** Column coordinate grid $x_{grid}$, disparity map $\tilde{d}_l$ |
| $Q \leftarrow x_{grid} - \tilde{d}_l$, $mask \leftarrow bool(Q > 0)$ |
| $Q_{down} \leftarrow floor(Q)$; $H, W \leftarrow shape(Q_{down})$ |
| **for** $i \in [0, W - 2)$ **do** |
|     $\hat{q} \leftarrow$ broadcast $Q_{down}[:, i]$ to $W - 1 - i$ columns |
|     update $mask[:, i]$ with $\min (bool(\hat{q} != Q_{down}[:, i: W - 1]), axis = 1)$ |
| **end** |
| **Output:** Occlusion map $mask$ |

| **Algorithm 3**: Domain adaptation—Stereo fine-tuning. |
|---|
| **Input:** Randomly shuffled $S_t(I_l, \tilde{I}_r, \tilde{d}_l)$, batch size $N$, max iteration $K$ |
| Initialize weights with pretrained weights |
| **repeat** |
|     **for** $k \in [0, len(S_t)/N - 1)$ **do:** |
|         feed $N$ samples to train |
|         $k \leftarrow k + 1$ |
|         **if** $k = K$ **then** break |
|     **end** |
| **until** $k = K$ |
| **Output:** An adapted stereo model |

and due to the absence of ground-truth disparities for infrared images, a suitable domain adaptation approach was proposed to generate precise training data while accommodating the characteristics of infrared images, as presented in Fig. 5. Using an off-the-shelf monocular depth estimation network, the depth maps were estimated for left images. Subsequently, fake disparity maps and corresponding right images were synthesized. These synthetic data were then used to fine-tune the stereo matching network, enabling the adaptation to infrared images of waves.

*B. Data Generation*

Supervised training of deep stereo networks requires rectified pairs of images $I_l$ and $I_r$, together with a ground truth disparity map $d_l$. Inspired by [37], [38], as described in Algorithm 1, the training dataset was synthesized for fine-tuning using collected stereo images. First, an off-the-shelf large model of monocular depth estimation network MiDaS [43] was used to estimate a plausible depth map $L$ for left image $I_l$, represented as $L = m_d(I_l)$. Although the depth estimation provided by the monocular network is often not highly accurate due to the scale effect, it could distinguish plausible near and far relationships. This capability was commonly employed for foreground and background segmentations of images. Subsequently, the disparity range $[d_{\min}, d_{\max}]$ of the infrared image dataset $D_t(I_l, I_r)$ collected in hydrodynamic experiments was estimated, and the depth map $L$ was mapped into a reasonable disparity map $\tilde{d}_l$. Subsequently, a fake right image $\tilde{d}_l$ was synthesized from the left image via forward warping. Finally, the missing pixels in the synthesized right image $\tilde{I}_r$ were sampled from the corresponding real right image $I_r$.

To ensure the quality of synthesized images in scenes with three-dimensional structures, the occluded regions were precomputed based on disparity map $\tilde{d}_l$, excluding pixels in the occluded regions in the warping process. Algorithm 2 presents an illustration of the algorithm for calculating occluded regions. To calculate the occlusions in a left image, $x_{grid}$ was the column coordinates of each pixel in the left image, subtracting $\tilde{d}_l$ from it yields $Q$, the target coordinates when reprojecting the left image pixels. After a Boolean operation for $Q$, values less than 0 indicated the regions beyond the field of view of the right image, resulting in a preliminary $mask$. Subsequently, pixelwise comparisons were made for each value of integer matrix $Q_{down}$. For each pixel, if an identical value existed on its right side at the same row, it was considered to be in the occluded region, and the $mask$ was subsequently updated.

Methods to generate data were employed for each left image to create a plausible tuple of training data $(I_l, \tilde{I}_r, \tilde{d}_l)$, ultimately forming a synthetic dataset. This dataset was then used for the domain adaptation training of the stereo neural network. Although the generated disparity $\tilde{d}_l$ did not represent actual disparity values, it served as ground truth relative to the input images $(I_l, \tilde{I}_r)$. Stereo matching was a generalized task, and the synthesized data were sufficient for the stereo neural network to adapt to the style of the target domain. Regarding the training dataset composition, 20–50 pairs of images were selected from each case (regular wave or wave–cylinder interaction), totaling 954 pairs of images.

*C. Fine-tuning the Stereo Network*

The stereo network (RAFT-Stereo) was then fine-tuned to adapt to the wave infrared image. As presented in Algorithm 3, the network was initialized with pretrained weights, and the synthesized data $S_t(I_l, \tilde{I}_r, \tilde{d}_l)$ were randomly shuffled and fed to the network for training. The fine-tuning details were as follows: batch size $N = 2$, maximum iteration steps $K = 20000$, and random cropping with a size of $320 \times 512$. The training duration of the network was measured in iteration steps, and a 0.5-px accuracy was taken as the training reference until convergence.

## IV. EXPERIMENTS FOR VALIDATION

To validate the feasibility and accuracy of the proposed approach using infrared stereo image sequences for the surface

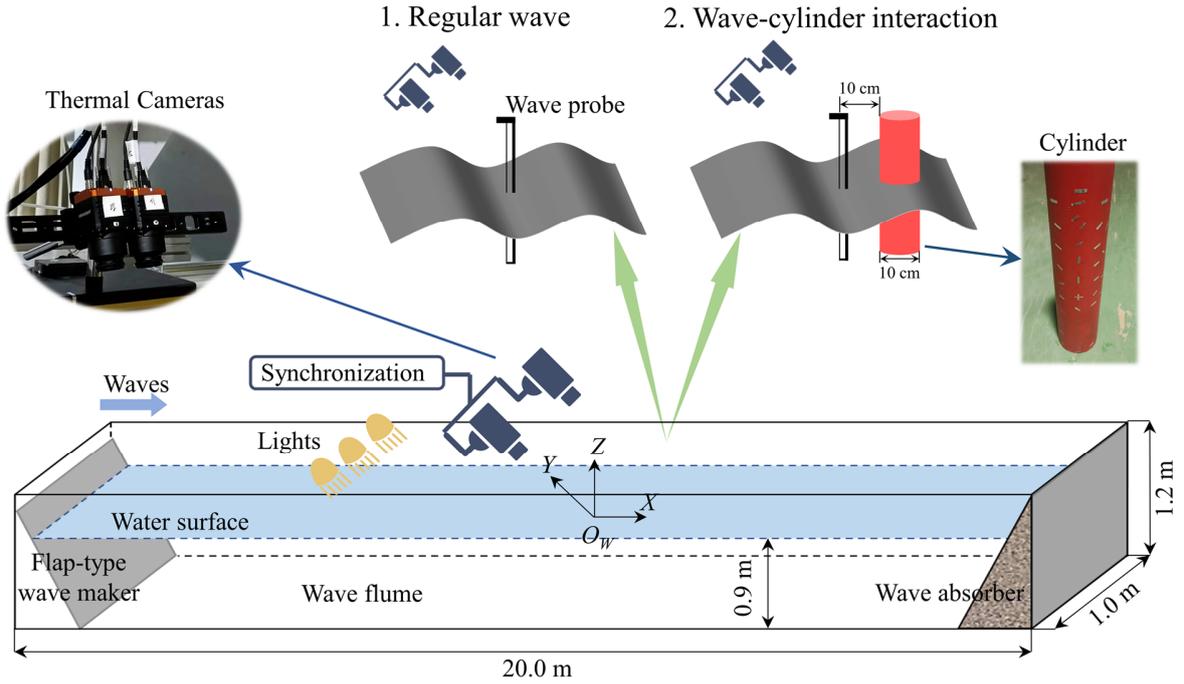

Fig. 6. Experimental layout in wave flume with two thermal cameras and three heating lights above the water surface.

TABLE I
PARAMETERS OF INFRARED CAMERAS

| Specifications | Parameters |
| --- | --- |
| Detector type | Vanadium oxide uncooled detector |
| Wavelength range | 8-14 μm |
| Pixel size | 17 μm |
| Resolution | 640×512 |
| Frame rate | 50 fps |
| Focal length | 12 mm |
| Field of view | 48.77°×39.87° |
| Thermal sensitivity | < 35 mK |

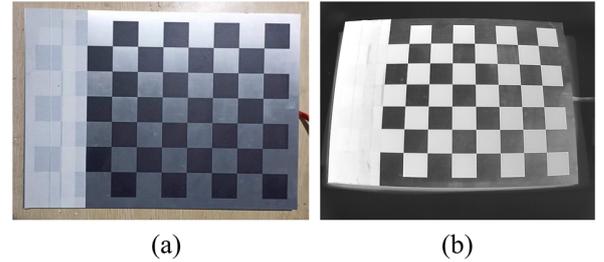

(a)       (b)

Fig. 7. Chessboard for geometric calibration: images at visible (a) and infrared (b) wavelengths.

reconstruction of water waves, several wave flume experiments were conducted.

*A. Experimental Layout*

As presented in Fig. 6, the wave flume had a length, width, and height of 20.0, 1.0, and 1.2 m, respectively. The water depth was 0.9 m. The waves were generated by a flap-type wave maker, and wave absorbers were installed at the opposite end.

Owing to the heat exchange between water and air, there is typically a cooler or warmer thin layer on the water surface. When the water surface is disturbed in waves, the fluid microstructures from the bottom tend to emerge at the water surface, forming the texture that can be seen in the infrared images. However, the mixing of fluids quickly reaches thermal equilibrium, resulting in the formation of textures that do not persist for an extended period. To enhance image contrast and ensure an adequate duration of texture presence, several heat lamps were arranged upstream in the measurement area to heat the surface water. The total power of the heat lamps was approximately 1 kW, and it was observed that they increased the surface water temperature in the measurement area by approximately 1°C–2 °C.

In the present study, the binocular setup consisted of two identical long-wave infrared cameras, and the main parameters of the cameras are listed in Table I. The baseline distance between the optical centers of the cameras was approximately 6 cm. The binocular cameras were positioned approximately 0.6 m above the water surface, with the optical axes inclined at an angle of 22.8° relative to the Z-axis. Time synchronization between the two cameras was achieved through multiprocess parallel triggering, with a synchronization error of less than 0.1 μs. The infrared cameras were equipped with online nonuniformity correction, which was performed after the cameras reached a stable temperature. Typically, a preheating period of approximately half an hour was required before imaging. Due to the individual variations among imaging units and the impact of device heating on imaging, shadow correction, and additional nonuniform denoising were necessary to process the acquired images.

TABLE II
PARAMETERS OF REGULAR WAVES

| $T$ (s) \ $H$ (cm) | $H/\lambda = 1/16$ | $H/\lambda = 2/25$ | $H/\lambda = 1/10$ |
|---|---|---|---|
| 0.553 | 3.09 | 4.05 | 5.23 |
| 0.632 | 4.04 | 5.28 | 6.82 |
| 0.712 | 5.10 | 6.67 | 8.60 |
| 0.791 | 6.29 | 8.22 | 10.58 |
| 0.949 | 9.03 | 11.77 | |
| 1.186 | 13.84 | | |

*B. Geometric Calibration*

Before image acquisition, accurate geometric calibration is necessary for image quantification, as well as epipolar and distortion rectifications. For thermal imaging cameras, the factors affecting image brightness include temperature and surface emissivity. Therefore, a heated aluminum-based chessboard calibration pattern was customized, similar to the black-and-white chessboard calibration patterns for traditional visible-light cameras, as presented in Fig. 7. Different emissivity coatings were applied to each grid to distinguish them, and the size of each square was 50 mm.

*C. Regular Wave Experiment*

Regular waves were generated and measured in the wave flume. Concurrent with the image acquisition, a wave probe was positioned on one side to obtain the time history of the wave elevations, which would be compared with the results of stereo reconstruction to evaluate the accuracy of the stereo method. The wave parameters are presented in Table II, where $H$ denotes the wave height; $T$, the wave period; and $H/\lambda$, the wave steepness. As an example, Figs. 8 (a) and (b) present the images captured during the propagation of regular waves with $H/\lambda = 2/25$ and $T = 0.533$ s. Due to the nonuniform heating upstream and significant Stokes drift induced by the large wave steepness, streak-like patterns were formed, as observed in the images.

*D. Wave–cylinder Interaction Experiment*

In the experiments of wave–structure interactions, wave scattering makes the wave field more complex. To evaluate the capability of the proposed method in measuring such complex wave fields, a fixed cylindrical column was positioned at the center of the wave flume to collect data on wave–cylinder interactions, and the incident wave parameters were the same as those in Table II. The cylinder surface was particularly painted to achieve an emissivity of approximately 0.95. Furthermore, to prevent stereo mismatching caused by the lack of texture on the cylinder surface, aluminum foil strips with a 0.05–mm thickness were randomly pasted on the surface, and the surface emissivity of the aluminum foil was 0.05–0.20. Therefore, even at the same temperature, the brightness in infrared imaging differs, facilitating a reasonable positioning for the cylindrical surface. The acquired images are presented in Figs. 8 (c) and (d), where the texture around the cylinder is sufficiently rich due to the wave scattering and breaking. Similarly, after the completion of image acquisition, a wave probe was positioned upstream, 10 cm in front of the cylinder, and the wave time histories were obtained under repeated wave cases for validation.

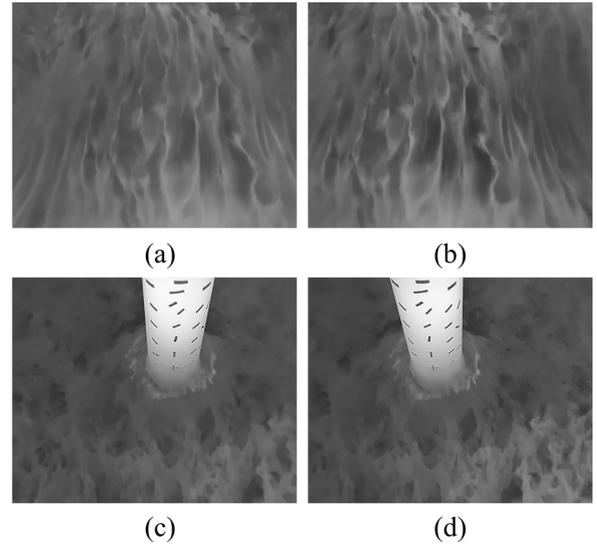

Fig. 8. Acquired infrared images (after rectifications): stereo image pair (a) and (b) in regular wave experiment, (c) and (d) in wave–cylinder interaction experiment.

V. RESULTS AND DISCUSSION

*A. Evaluation of Disparity*

The fine-tuned model was tested on the testing dataset. Regarding the composition of the testing dataset, 10-20 pairs of images not present in the training dataset were selected from each wave case, totaling 350 pairs. The fine-tuned disparity maps are compared with the results obtained using the traditional semi-global block matching (SGBM) algorithm and the pretrained model of RAFT-Stereo, as presented in Fig. 9.

It can be clearly observed that the SGBM algorithm poorly performed on the infrared images. The disparity maps exhibit many holes and mismatches, with particularly deficient performance in the matching of the cylindrical surface and estimation of disparities in the occluded region. This can be attributed to the fact that the brightness of thermal images mainly depends on the intensity of infrared radiation emitted outward from the object. As every object serves as a radiation source, there are no shadows or variations in brightness caused by ambient light. Consequently, the stereo matching becomes challenging. In addition, the traditional algorithm is sensitive to the image noise, further affecting its performance.

Although the pretrained model of RAFT-Stereo has exhibited improved performance, it still suffers from instability, occasional mismatches, random errors, and suboptimal handling of details, particularly in the region where the cylinder meets the water surface—an area of particular interest in hydrodynamic research. The fine-tuned model has effectively addressed these issues. The transition at the interface between the cylinder and water is now more reasonable, random errors are eliminated, and the disparity is continuous across the cylindrical surface.

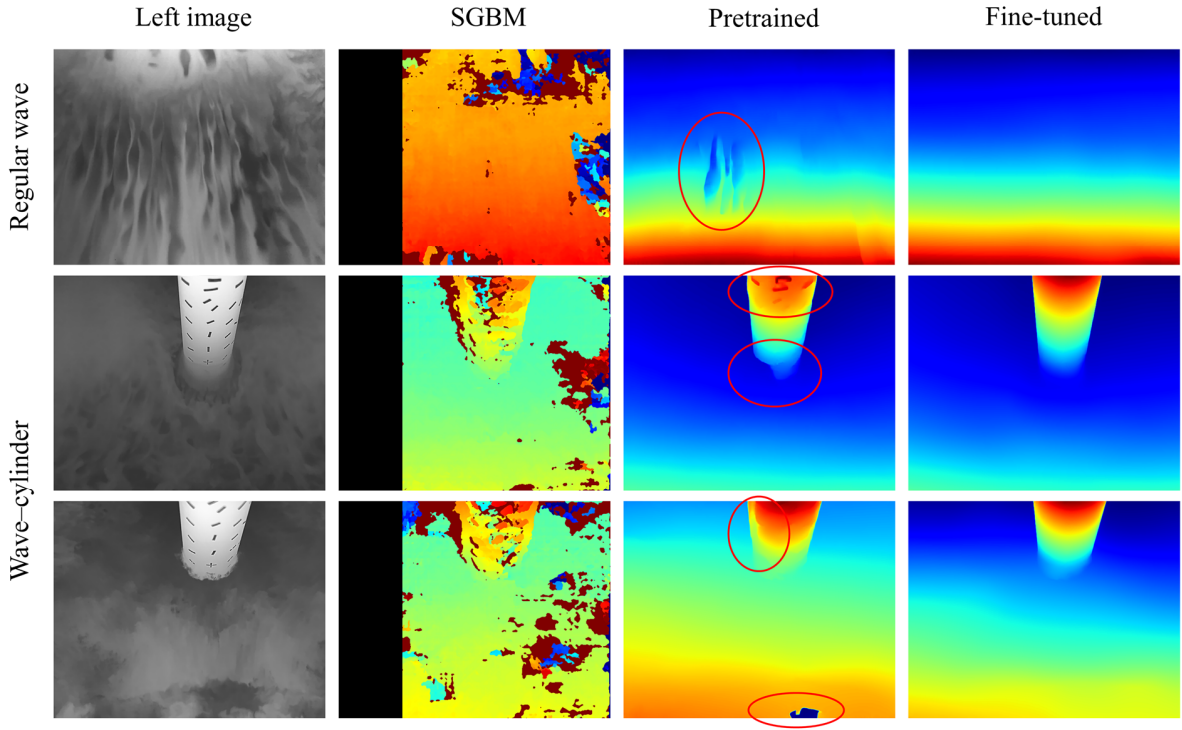

Fig. 9. Qualitative results of disparity map. From left to right: input left image and disparity maps generated by SGBM, pretrained RAFT-Stereo model on SceneFlow [31], and fine-tuned RAFT-Stereo model.

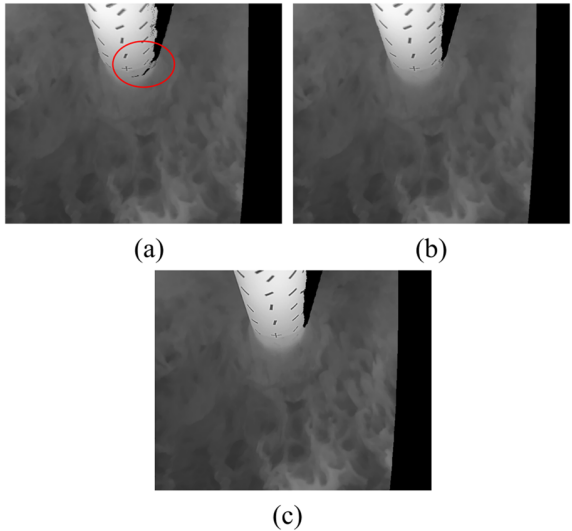

(a)      (b)

(c)

Fig. 10. Comparison of right images: (a) reprojected by pretrained model, (b) reprojected by fine-tuned model, and (c) ground truth.

TABLE III
EVALUATION METRICS

| Model | SSIM | PSNR | MSE | HD |
|---|---|---|---|---|
| Pretrained | 0.9677 | 27.0331 | 155.0628 | 3.8977 |
| Fine-tuned | **0.9754** | **28.2597** | **121.4140** | **1.0933** |

Note: for SSIM and PSNR, higher is better; for MSE and HD: lower is better.

To further quantify the accuracy of the obtained disparity, inspired by self-supervised research [29], the photometric reprojection error was employed to measure model performance before and after fine-tuning. The left image was projected into the right image using the disparity map, and the projected right image was then compared with the real right image, as presented in Fig. 10. Such a comparison was an indirect method to examine the performance in that the disparity error was less than 1 px. Notably, the pixel values in occluded regions cannot be obtained, thus the evaluation results are presented only for nonoccluded regions. The metrics used in the comparison include structural similarity (SSIM) [44], mean squared error (MSE), peak signal-to-noise ratio (PSNR) [45], and Hausdorff distance (HD) [46]. SSIM assesses the overall similarity between two images. MSE and PSNR measure the statistical errors of the images, and HD evaluates the similarity of image boundary features, providing an indirect evaluation of disparity estimation in occluded regions. Table III presents the performance of the pretrained and fine-tuned models on these metrics, indicating an overall improvement in accuracy after fine-tuning.

*B. 3D Reconstruction Performance*

The most significant advantage of the stereo method is its ability to obtain a 3D representation of the wave field. After obtaining the disparity maps, 3D point clouds were reconstructed in the camera coordinate system for each frame. For the present application, no ground truth for the wave surface was available for cross-validation. The time histories of elevation acquired by wave probes were limited at single

TIM-24-00463
2

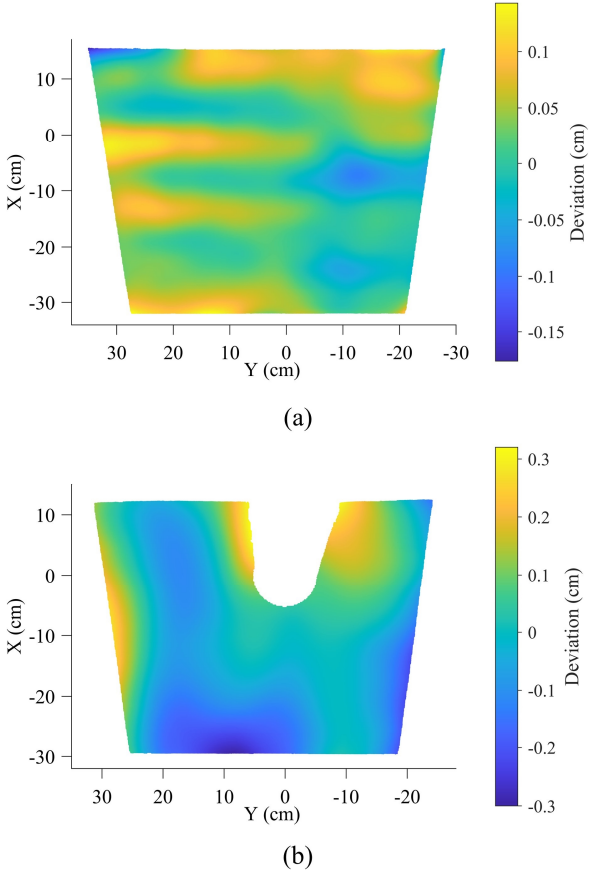

Fig. 11 Deviation maps for reconstructed regions: (a) regular wave experiment and (b) wave-cylinder interaction experiment.

points. Instead, a stereo image sequence of a flat-water surface was acquired to serve as ground truth. For this image sequence, it is known that the imaged surface was flat. Therefore, the reconstructed 3D points should lie in a plane. If the position and orientation of this plane are known with high accuracy, the distance of the reconstructed points from this plane can be used to evaluate metrological 3D reconstruction performance.

A plane was robustly fitted to the reconstructed points using the random sample consensus (RANSAC) method. The so-obtained fit plane was then used as a reference against which the points reconstructed from single frames were compared, and the deviation map is shown in Fig. 11. Overall, the metrological bias of the 3D reconstruction of the still water surface is small, both in the millimeter scale. In the wave-cylinder interaction experiments, due to a somewhat more complex scene, the larger deviation occurs at edges and in occluded regions, and the smaller deviation at the center. The standard deviations in the reconstructed regions are 0.03 cm and 0.11 cm, respectively.

Subsequently, a world coordinate system was defined based on the reconstruction of the still water surface, and all point clouds were transformed into this system, as shown in Fig. 12. It can be seen that the shape of the regular wave field was smooth and continuous, consistent with the empirical knowledge. In the presence of a cylinder, the complex wave field generated by wave scattering was also reconstructed successfully. To address the wave run-up phenomenon, particular attention was given to the maximum wave run-up height. Using the proposed method, the 3D wave field near the intersection of the cylinder and water surface can be reconstructed, as shown in Fig. 13. Compared with the wave probe, the proposed method provided a more accurate determination of the maximum wave run-up height and the spatial distribution of the wave field near the structure, which contributes to further studies of wave-cylinder interaction.

One pair of cameras was used in this study, resulting in the reconstructed wave fields limited to the upstream side of the cylinder, affected by occlusion. To capture the complete evolution of waves around the structure, the use of an additional pair of cameras downstream is suggested. Both pairs of cameras are required to be synchronized in time, and two reconstructed results can be fused through the coordinate transformation.

*C. Evaluation of Time Series*

From the point cloud sequence by stereo reconstruction of the infrared image sequence, the time histories of wave elevations were extracted and compared with those obtained through the wave probe in experiments. The wave probe is linear-measuring instrument, and its coefficient underwent multiple calibrations during experiments. The average of the calibration coefficients is adopted, with a standard deviation of 0.44%. Therefore, the measurements by wave probe were taken as the target to evaluate the results of stereo imaging method.

Figs 14 (a)–(d) present the comparisons of regular wave time histories, showing excellent agreements between the measurements. To quantify and evaluate the accuracy of the stereo method, a randomly sampled scatter plot is presented in Fig. 14 (e). Taking the measurements of the wave probe as the target, the linear fit closer to the identity map $y = x$ indicates better overall measurement by stereo imaging. In Fig. 14 (e), the slope of the linear fit curve is 0.979, indicating a mean bias of approximately 2.1% for the stereo method in the measurement of regular waves. Similarly, Figs 15 (a)–(d) present the comparisons of time histories in wave–cylinder interactions, with a mean bias of approximately 1.5% shown in Fig. 15 (e). The measurement accuracy of the proposed method is demonstrated to be sufficiently high, allowing for accurate spatiotemporal measurement of the wave field.

In addition, to further validate the accuracy of stereo measurements, the statistics of wave histories are listed in Table IV and Table V, including the mean wave height $\bar{H}$, mean period $\bar{T}$, and R-squared of two curves. The definition of R-squared is described as:

$$R^2 = 1 - \frac{\sum_{i=1}^{N}(y_i^p - y_i^s)^2}{\sum_{i=1}^{N}(y_i^p - \bar{y}^p)^2} \quad (4)$$

where $y_i^s$ is the wave elevation measured by stereo method, $y_i^p$ is the wave elevation measured by wave probe and $\bar{y}^p$ is the mean value of all $y_i^p$.

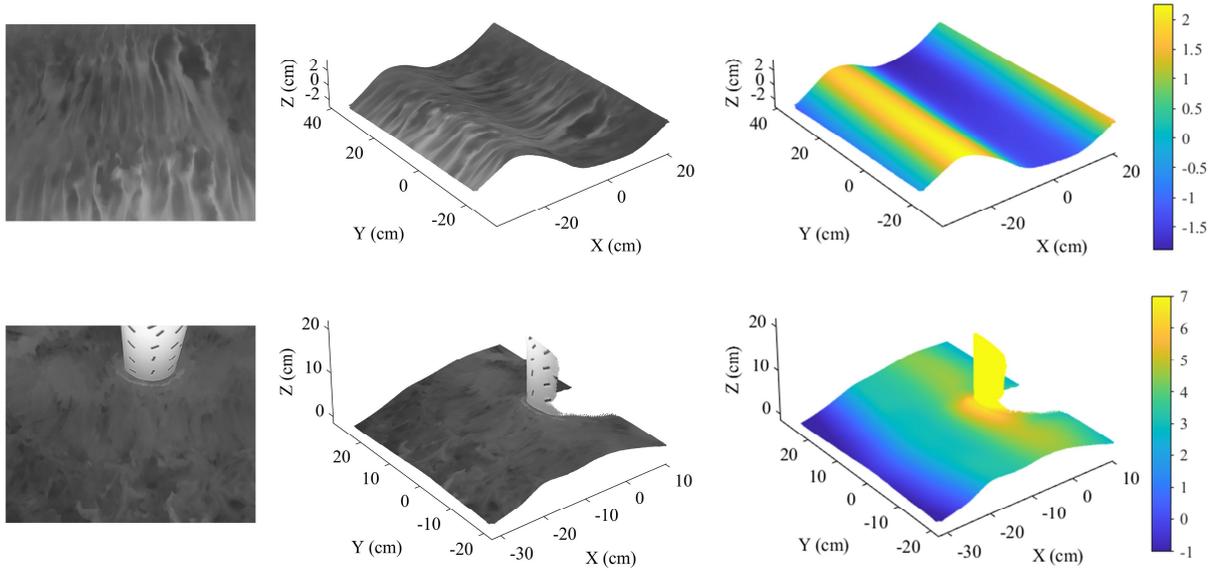

Fig. 12. 3D reconstruction results. From left to right: input left image, reconstructed point cloud with natural texture, and point cloud with pseudo-color.

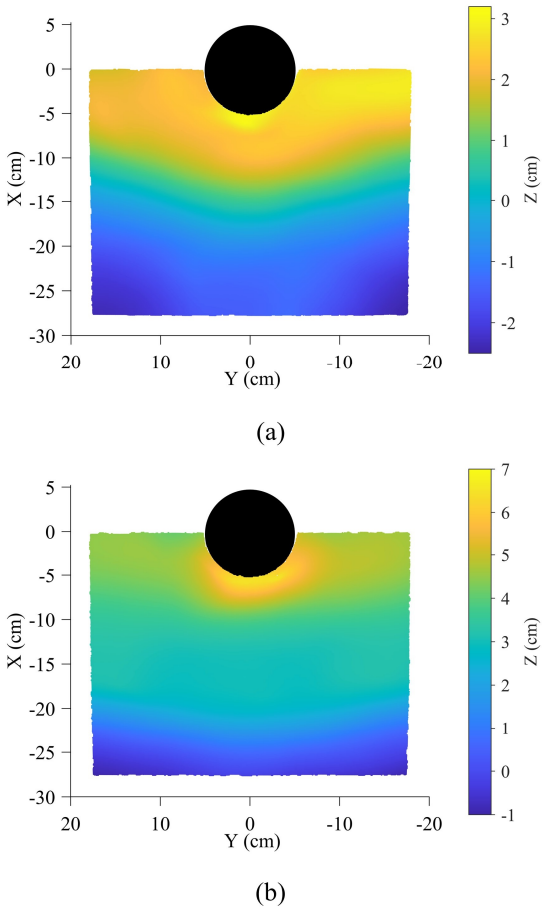

Fig. 13 Close-ups of the near-field wave runup when wave crests are reaching the cylinder: (a) $H/\lambda = 2/25$, $T = 0.553$ s, (b) $H/\lambda = 2/25$, $T = 0.791$ s

The statistics show that the estimation of wave period is easier due to the camera's internal clock frequency being sufficiently accurate. The $R$-squared shows a great agreement between the two curves. The measurement error of wave height is slightly larger, which originates from many factors, including image quantization, calibration, and mismatching. The calibration error and mismatching error can be effectively avoided by all means. The quantification error is an intrinsic error that represents the quantitative measurement capability of the binocular stereo system.

*D. Quantification Analysis*

The quantization error is inherent in stereo reconstruction, arising from the discretization of the digital image. Although the measurement accuracy is sufficiently high, it is helpful to delve into the reasons affecting the quantification error for exploring the most suitable camera arrangement in other applications.

The positioning error $e$ of a pixel is commonly assumed to be less than a pixel size, and the disparity error is usually estimated as $e_d = e/\sqrt{2}$ [8]. Through differentiation of $z = B \cdot f/d$, the quantification errors $e_z$, $e_x$, and $e_y$ along the $z$, $x$, and $y$-axes in the camera coordinate system, respectively, are given as follows:

$$e_z = \frac{z^2}{\sqrt{2}B \cdot f} e \qquad (5)$$

$$e_x = \sqrt{1 + \left(\frac{x'}{\sqrt{2}B}\right)^2} \frac{z}{f} e \qquad (6)$$

$$e_y = \sqrt{1 + \left(\frac{y'}{\sqrt{2}B}\right)^2} \frac{z}{f} e \qquad (7)$$

The ultimate expected outcome is in the world coordinate system, and the errors in the three directions are represented by the following:

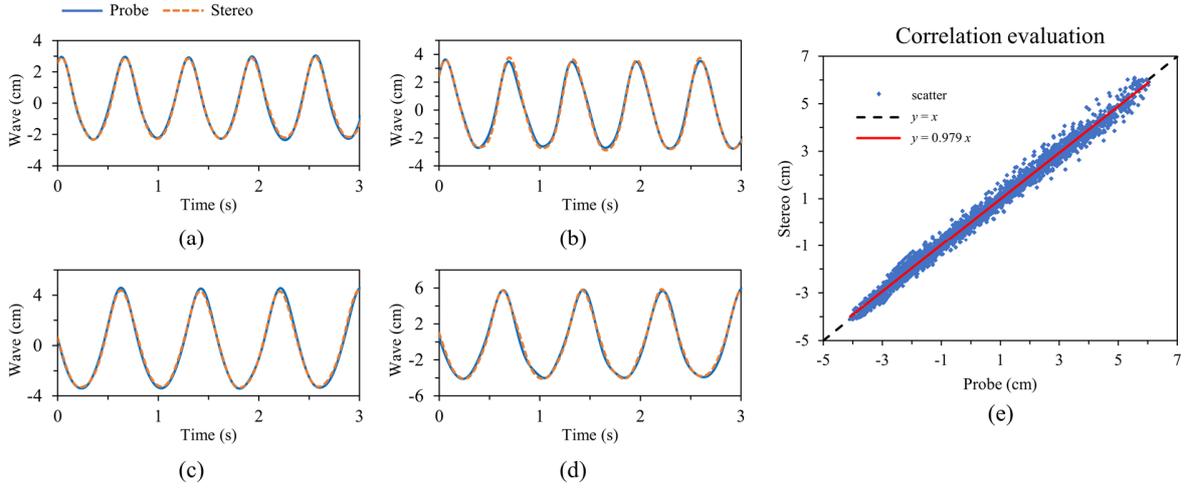

Fig. 14. Comparisons of regular wave time histories: (a) $H/\lambda = 2/25$, $T = 0.632$ s, (b) $H/\lambda = 1/10$, $T = 0.632$ s, (c) $H/\lambda = 2/25$, $T = 0.791$ s, and (d) $H/\lambda = 1/10$, $T = 0.791$ s. (e) is the scatter plot evaluating the correlation between the results measured using the stereo imaging method and the wave probe: linear fit (solid red line) and perfect fit (black dashed line).

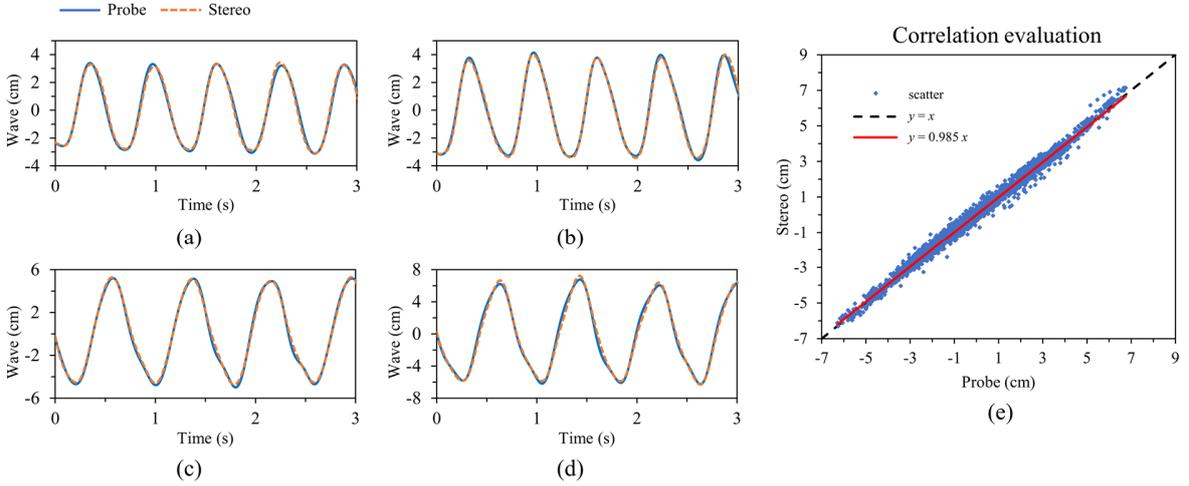

Fig. 15. Comparisons of wave time histories in wave–cylinder interactions: (a) $H/\lambda = 2/25$, $T = 0.632$ s, (b) $H/\lambda = 1/10$, $T = 0.632$ s, (c) $H/\lambda = 2/25$, $T = 0.791$ s, and (d) $H/\lambda = 1/10$, $T = 0.791$ s. (e) is the scatter plot evaluating the correlation between the results measured using the stereo imaging method and the wave probe: linear fit (solid red line) and perfect fit (black dashed line).

TABLE IV
STATISTICAL COMPARISON OF REGULAR WAVES

| Waves | $\bar{H}$ (cm) | | Error | $\bar{T}$ (s) | | Error | $R^2$ |
|---|---|---|---|---|---|---|---|
| | Stereo | Probe | | Stereo | Probe | | |
| $H/\lambda = 2/25$, $T = 0.632$ s | 5.32 | 5.3 | 0.38% | 0.6312 | 0.6325 | 0.21% | 99.10% |
| $H/\lambda = 1/10$, $T = 0.632$ s | 6.36 | 6.18 | 2.91% | 0.6328 | 0.6333 | 0.08% | 99.34% |
| $H/\lambda = 2/25$, $T = 0.791$ s | 7.91 | 7.81 | 1.28% | 0.79 | 0.7909 | 0.11% | 98.96% |
| $H/\lambda = 1/10$, $T = 0.791$ s | 9.79 | 9.83 | 0.41% | 0.791 | 0.7905 | 0.06% | 99.17% |

TABLE V
STATISTICAL COMPARISON OF WAVES IN WAVE-CYLINDER INTERACTION

| Waves | $\bar{H}$ (cm) | | Error | $\bar{T}$ (s) | | Error | $R^2$ |
|---|---|---|---|---|---|---|---|
| | Stereo | Probe | | Stereo | Probe | | |
| $H/\lambda = 2/25$, $T = 0.632$ s | 6.04 | 6.16 | 1.95% | 0.631 | 0.6361 | 0.80% | 98.99% |
| $H/\lambda = 1/10$, $T = 0.632$ s | 7.21 | 7.33 | 1.64% | 0.6335 | 0.6357 | 0.35% | 99.24% |
| $H/\lambda = 2/25$, $T = 0.791$ s | 9.37 | 9.47 | 1.06% | 0.7902 | 0.791 | 0.10% | 99.53% |
| $H/\lambda = 1/10$, $T = 0.791$ s | 12.66 | 12.47 | 1.52% | 0.7912 | 0.7916 | 0.05% | 99.00% |

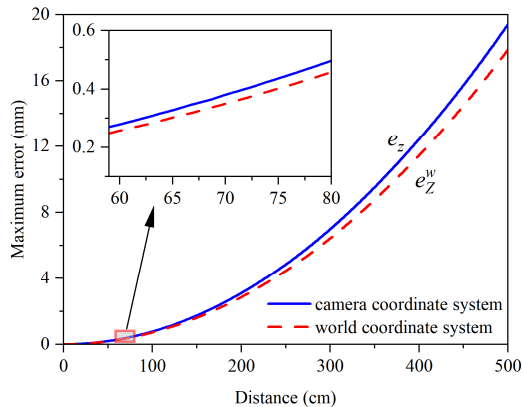

Fig. 16. Maximum values of quantification errors in camera ($e_z$) and world ($e_Z^w$) coordinate systems.

$$\begin{bmatrix} e_X^w \\ e_Y^w \\ e_Z^w \end{bmatrix} = \mathbf{R} \begin{bmatrix} e_x \\ e_y \\ e_z \end{bmatrix} \quad (8)$$

where **R** denotes the rotation matrix that transforms the camera coordinate system to the world coordinate system.

Assuming that the positioning error $e$ takes the maximum value of 1 pixel, the maximum value of quantification error $e_z$ was calculated using the camera parameters and shown in Fig. 16, where the magnified portion corresponds to the working range of cameras in experiments. $e_z$ and $e_Z^w$ are proportional to the square of the distance between the camera and the measured object. It is important to control the distance-to-base ratio $z/B$ and image scale $z/f$. Although ensuring that the overlapping field of view meets the requirements, the distance should be sufficiently short to reduce inherent error. Furthermore, the camera tilt angle should be carefully managed to avoid reflections and should not be excessively inclined, as this could lead to distortion in distant areas.

The limitations of the reconstruction strategy are also summarized. The domain adaption mainly addresses potential random errors and mismatches in image details, aiming to output a stable disparity map. The overall matching quality is predominantly determined by the image quality and the performance of the network. Further improvements, such as the use of higher-power laser heating for the water surface, will contribute to obtainment of images with rich textures. Moreover, the improvement of stereo network will facilitate the enhancement of the accuracy and efficiency of stereo matching. In addition, the disparity estimation in occluded regions cannot be effectively validated, and an improved evaluation metric is expected.

## VI. CONCLUSIONS

In this study, a novel technique was proposed for measuring free water surface, including two components: data acquisition based on thermal stereography and a reconstruction strategy involving deep neural networks. This technique was developed for the continuous and accurate 3D measurement of wave fields using binocular stereo imaging in indoor facilities, without the need for any markers on the water surface during image acquisition. The monocular depth estimation and stereo matching networks used in the reconstruction process can also be replaced with other advanced models, leveraging the advancements in deep learning for greater potential in terms of accuracy and efficiency for this technique.

This study validated the efficacy and feasibility of the proposed technique in the spatiotemporal reconstruction of wave fields and demonstrated high accuracy with a mean bias of less than 2.1% compared with the traditional method. The proposed technique can serve as an effective tool for studying wave evolutions in hydrodynamic experiments.

ACKNOWLEDGMENT

The authors would like to thank the anonymous reviewers and editors for their profound comments.

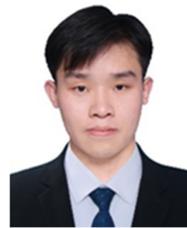
**Deyu Li** received the bachelor's degree in naval architecture and ocean engineering from Harbin Engineering University, Harbin, China, in 2020. He is currently pursuing the Ph.D. degree with the School of Naval Architecture, Ocean & Civil Engineering, Shanghai Jiao Tong University, Shanghai, China.

His research focuses on spatiotemporal measurement of wave fields based on computer vision.

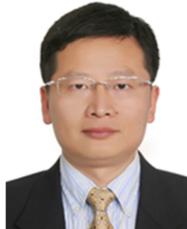
**Longfei Xiao** is Tenure Professor at School of Naval Architecture, Ocean and Civil Engineering in Shanghai Jiao Tong University, China. He has worked in the State Key Laboratory of Ocean Engineering since 1998, and as a Senior Visiting Scholar at Newcastle University in UK from 2013 to 2014.

The scope of research work consists of hydrodynamics of floating offshore structures with focus on physical and numerical modelling of environments, motions and loads of floating platforms, deepsea mining system, and fluid-structure interactions. He has completed more than 60 research projects from National Natural Science Foundation of China, and etc. In recent five years, he has published more than 80 papers, including 51 SCI articles.

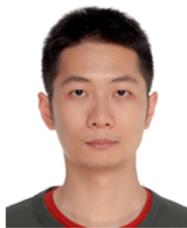
**Handi Wei** (Member, IEEE) received the Ph.D. degree from Shanghai Jiao Tong University, Shanghai, China, in 2019.

He is currently a research fellow with the State Key Laboratory of Ocean Engineering, Department of Civil and Ocean Engineering, Shanghai Jiao Tong University. His work focuses on hydrodynamic analysis and environmental sensing of offshore and underwater equipment.

TIM-24-00463

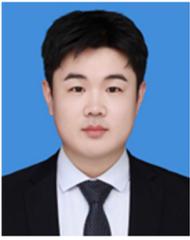

**Yan Li** received the bachelor's degree in naval architecture and ocean engineering from Harbin Engineering University, Harbin, China, in 2018. He is currently pursuing the Ph.D. degree with the School of Naval Architecture, Ocean & Civil Engineering, Shanghai Jiao Tong University, Shanghai, China.

His research focuses on nonlinear hydrodynamic responses simulation of offshore platforms based on deep learning.

**Binghua Zhang** received the MBA from Manchester Business School, MSc. from Norwegian University of Science and Technology, Trondheim, Norway, and bachelor's degree from Shanghai Jiao Tong University, Shanghai, China.

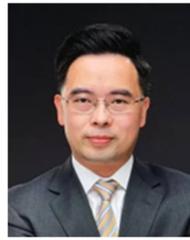

He has decades of experience in maritime industry, featuring strategic planning for Lloyd's Register global maritime businesses, Chairman of Maritime & Finance Excellence Center in Shanghai, President of Maritime London in China, and founding AI enabling autonomous shipping technology company Marautec Co. Ltd and Shipping AI Data Center. He was awarded Top 10 figureheads in China's Shipping Technologies in 2019 and in 2022, and Top 100 Most Influential Persons in China's Shipping Industries in 2022.